\documentclass[letterpaper, 10 pt, conference]{ieeeconf}  
\usepackage{times}

\IEEEoverridecommandlockouts    

\usepackage{times} 
\usepackage{graphicx} 
\usepackage[utf8]{inputenc} 
\usepackage{setspace} 
\usepackage{hyperref} 
\usepackage[noadjust]{cite}
\usepackage{flushend} 
\usepackage{booktabs}

\hypersetup{bookmarksopen,bookmarksnumbered,
pdfpagemode=UseOutlines,
colorlinks=true,
linkcolor=blue,
anchorcolor=blue,
citecolor=blue,
filecolor=blue,
menucolor=blue,
urlcolor=blue
}

\usepackage[font={small}]{caption}
\usepackage{subcaption}

\usepackage{amsmath,amsfonts,amssymb,amsopn,bm}

\newcommand{\E}{\mathbb{E}}
\newcommand{\R}{\mathbb{R}}

\usepackage[linesnumbered,ruled,vlined]{algorithm2e}  

\SetCommentSty{mycommfont}
\SetKwInOut{Input}{Input}                
\SetKwInOut{Output}{Output}              
\SetKwInOut{Requires}{Requires}
\SetKwInOut{Note}{Note}
\SetKwProg{Fn}{Function}{ is}{end}

\newcommand{\norm}[1]{\left\|#1\right\|}

\newcommand{\twonormsq}[1]{\|#1\|_2^2}

\newcommand{\xxnote}[3]{}
\ifx\hidenotes\undefined
  \usepackage{color}
  \renewcommand{\xxnote}[3]{\color{#2}{#1: #3}}
\fi

\newcommand\figref{Fig.~\ref}

\usepackage{paralist}
\usepackage{microtype}

\title{\LARGE \bf
Online augmentation of learned grasp sequence policies \\
for more adaptable and data-efficient in-hand manipulation
}

\author{Ethan K. Gordon$^{1}$,
Rana Soltani Zarrin$^{2}$
\thanks{$^{1}$ Ethan K. Gordon is with the Department of Computer Science and Engineering, University of Washington, Seattle, WA 98195
        {\tt\small ekgordon,@cs.washington.edu}. Work done as an intern at the Honda Research Institute.}
\thanks{$^{2}$ Rana Soltani Zarrin is with the Honda Research Institute, San Jose, CA 95134
        {\tt\small rana\_soltanizarrin@honda-ri.com}}
}

\begin{document}

\maketitle
\thispagestyle{empty}
\pagestyle{empty}

\IEEEpeerreviewmaketitle
\begin{abstract}
When using a tool, the grasps used for picking it up, reposing, and holding it in a suitable pose for the desired task could be distinct. Therefore, a key challenge for autonomous in-hand tool manipulation is finding a sequence of grasps that facilitates every step of the tool use process while continuously maintaining force closure and stability. Due to the complexity of modeling the contact dynamics, reinforcement learning (RL) techniques can provide a solution in this continuous space subject to highly parameterized physical models. However, these techniques impose a trade-off in adaptability and data efficiency. At test time the tool properties, desired trajectory, and desired application forces could differ substantially from training scenarios. Adapting to this necessitates more data or computationally expensive online policy updates. 

In this work, we apply the principles of discrete dynamic programming (DP) to augment RL performance with domain knowledge. Specifically, we first design a computationally simple approximation of our environment. We then demonstrate in physical simulation that performing tree searches (i.e., \emph{lookaheads}) and policy \emph{rollouts} with this approximation can improve an RL-derived grasp sequence policy with minimal additional online computation. Additionally, we show that pretraining a deep RL network with the DP-derived solution to the discretized problem can speed up policy training.
\end{abstract}

\section{Introduction}
Dexterous manipulation focuses on the problem of enabling stable grasping and manipulation of objects using multi-fingered hands~\cite{okamura_overview_2000}. In this broad space, many previous studies address object reposing only~\cite{van2015learning,kumar2016optimalControlRL,li_learning_2019}. In this work, we focus on the specific application of in-hand tool manipulation, which requires the object (i.e., the \emph{tool}) to be held with a grasp suitable for applying an arbitrary wrench (i.e., force and torque) to the external environment.

\begin{figure}[t!]
    \centering
    \includegraphics[width=\linewidth]{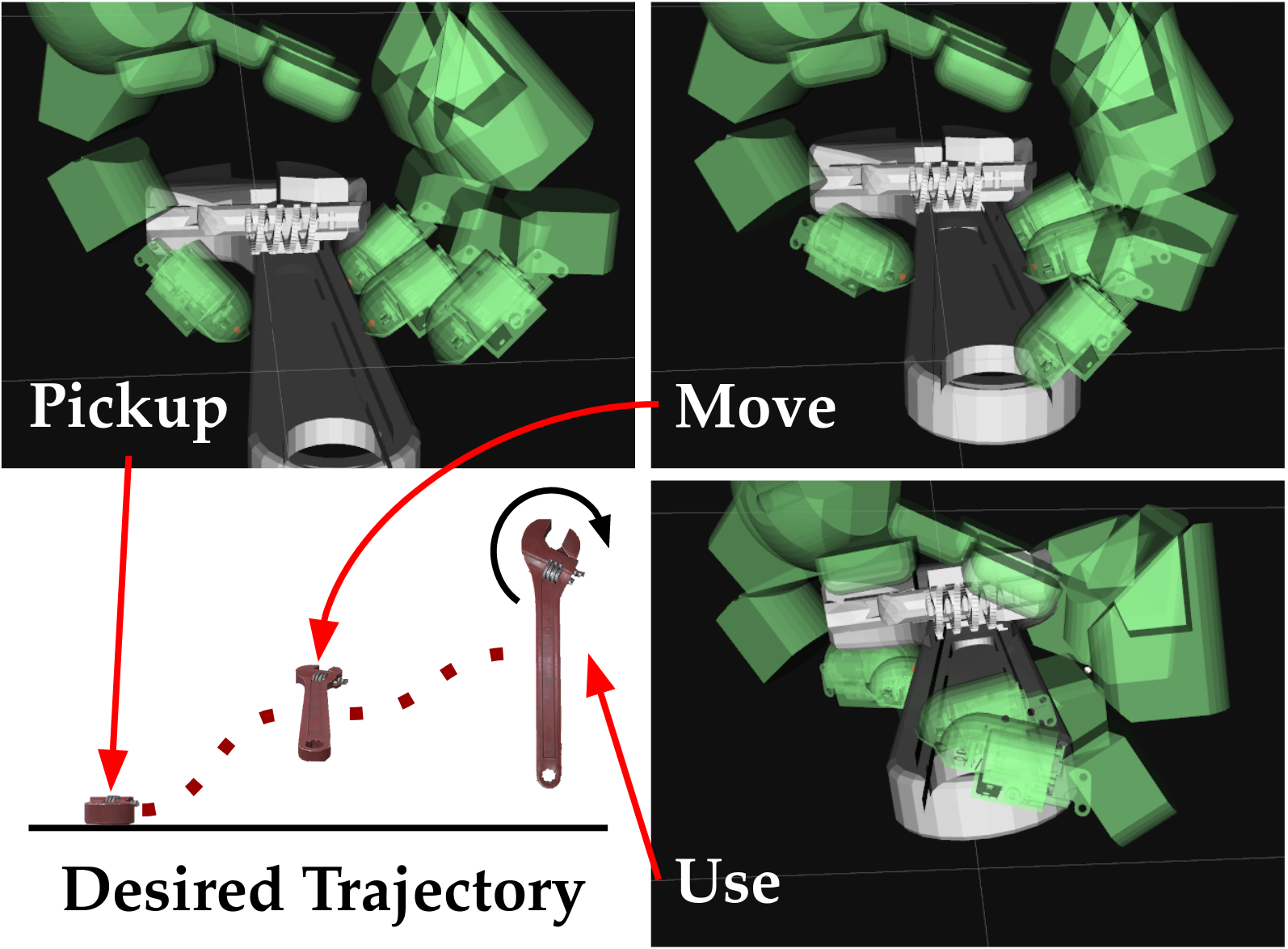}
    \caption{Given a desired tool trajectory and anticipated external forces, determine a sequence of grasps that continuously maintains force closure and stability. In this example, while moving the wrench, the selected grasp adds the ring finger for stability. To use the wrench for bolt tightening, the fingers gradually transition to a power grasp while the thumb is placed towards the wrench head to apply the desired torque.}
    \label{fig:intro}
\end{figure}

As shown in \figref{fig:intro}, this may require multiple different grasps addressing the sub-problems of acquisition, movement, and use. Therefore, we consider in-hand tool manipulation as a problem of \emph{grasp sequence planning}. Given a desired tool trajectory and application, the robot must determine which grasp to use at each point in time to execute the motion while maintaining stability over the entire episode. This is a difficult problem for traditional physical controllers. The dynamics are complex and highly parameterized. Future grasp performance depends on previous grasps. Prerequisite physical models of the system may be imperfect.

Deep reinforcement learning (DRL) offers an enticing solution by eliminating the need to model the complex physics of the problem ~\cite{yu_dexterous_2022}. End-to-end systems collect hours of data to directly learn low-level control policies~\cite{andrychowicz2020OpenAI_dexterousManipulation}. Hierarchical frameworks, on the other hand, can use RL to learn high-level sub-tasks or primitives and rely on model-based methods for lower level control~\cite{nasiriany2022primitives,kroemer2021learning_review}. Without explicit domain knowledge, both of these approaches effectively build physical models from scratch. Therefore, with insufficient data, these systems can struggle to adapt to scenarios that differ significantly from those at training time.

Our key insight is that discrete-time DP can be used to introduce domain knowledge to an RL policy at test time via a computationally-inexpensive approximate physical model. This scheme improves RL performance on previously unseen scenarios without additional data collection or computational overhead at test time. In summary, this work presents two primary contributions: (1) a surrogate environment model that can approximate the grasp sequence planning problem, and (2) a scheme that leverages traditional DP to augment test time RL performance using this approximate model. Additionally, we show that DP-derived grasp sequences within the surrogate model can be used as an expert policy to provide a ``hot start'' to RL training via behavior cloning. Finally, we compare this scheme to RL and DP baselines in physical simulation.

\begin{figure*}[t!]
    \centering
    \includegraphics[width=\textwidth]{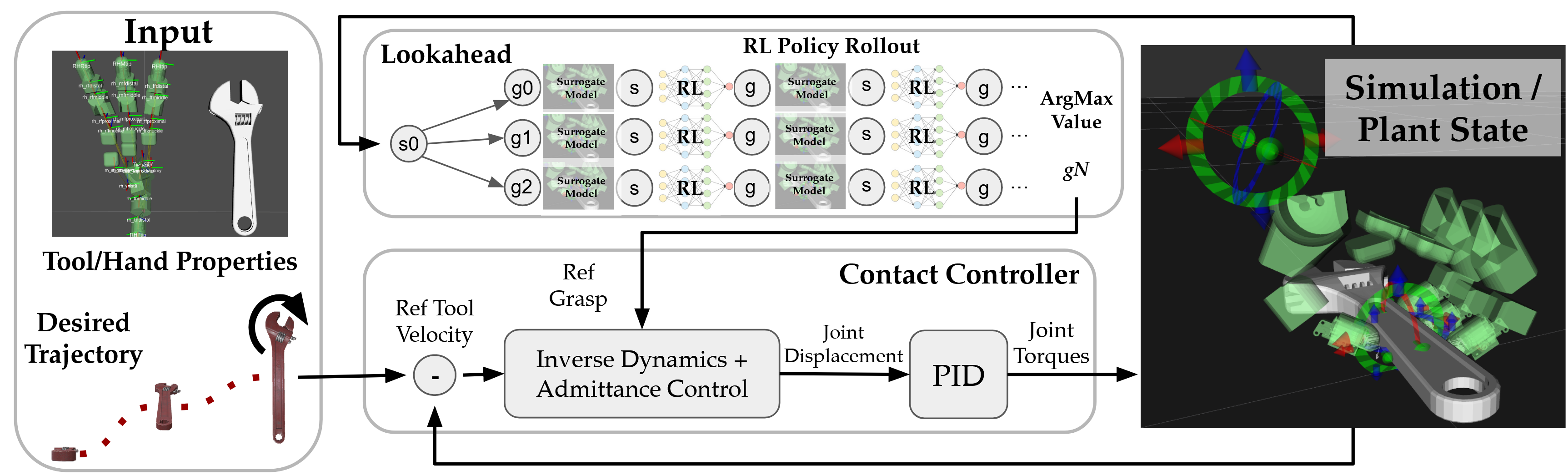}
    \caption{System diagram. At each time step, we construct a ``lookahead'' complete search tree of length $l$ for the discrete set of possible actions (i.e., grasps) from the current system state. The outcome of each grasp is approximated by the physical model described in Section \ref{sec:environment}. The terminating cost of each tree leaf is determined by rolling out the RL policy until the end of the episode. The grasp action with the highest value is set as the reference grasp for the Contact Controller from \cite{zarrin_hybrid_2022}, which in turn calculates the desired joint torques. }
    \label{fig:system}
\end{figure*}
\section{Related Work}
\label{sec:prev}
\paragraph{In-Hand Manipulation} While prior strategies for in-hand manipulation commonly use external or contact forces from the environment---such as supporting an object by the hand's palm, using non-prehensile primitives~\cite{dafle2014extrinsic,chavan2015prehensile,karayiannidis2015hand,van2015learning,nonprehensile_liu}, or using underactuated hands to improve stability and robustness~\cite{liarokapis2017deriving,abondance2020dexterous, van2015learning}---we perform in-hand manipulation considering internal forces using only fully actuated hands.
Model-based approaches have been used to perform in-hand manipulation tasks involving sliding and rolling~\cite{liu2009dextrous,mordatch_CIO} in computer graphics and simple object reorientation applications \cite{nonprehensile_liu};  however, the difficulty of modeling complex contact interactions and inaccuracies, and the high computation cost of re-solving the problem for variations introduced during run-time, make them unsuitable for real-world robotic applications.
As a result, data-driven approaches have been widely explored, albeit mainly for object reorientation. Taking advantage of the extra stability that results from supporting the object using the palm, multiple end-to-end DRL methods have been used for in-hand reorientation of simple objects, such as a cylinder or cube, using vision sensory information \cite{kumar2016optimalControlRL,andrychowicz2020OpenAI_dexterousManipulation}. However, data dependencies limit the sample efficiency and generalizability of these approaches. 

Hierarchical frameworks, i.e., those that learn motion primitives and rely on model-based low-level controllers for primitive execution, can alleviate the issues associated with individual methods.~\cite{li_learning_2019} has proposed a hierarchical learning/control scheme, where a higher level learning algorithm plans a sequence of motion primitives (reposing, flipping, sliding) that are executed by a lower level controller. In our recent work \cite{zarrin_hybrid_2022}, we developed a hybrid model- and RL-based grasp sequence planner capable of commanding object reposing and adding/removing/sliding contacts that are executed by a low-level controller; we showed higher robustness to trajectory variations compared to a pre-computed optimal sequence generated by solving a DP. In this paper, by pretraining RL with a DP-generated policy for a nominal environment, we speed up training, and by using an  online lookahead and rollout method, we make the planner more robust to trajectory and object variations at runtime.

\paragraph{Search-Based Lookahead}
The benefit of using search-based methods, a staple of traditional DP, to augment the performance of trained-RL in naturally discrete board game systems (such as the AlphaZero from DeepMind \cite{silver_mastering_2017} to solve Go) has been demonstrated. These methods were partially formalized in \cite{bertsekas_lessons_2021}, identifying the lookahead and rollout procedure. It contemplates the potential use of the idea in continuous, adaptive control but offers no specific implementation. Our work leverages this idea of augmenting RL with search-based DP: by discretizing the continuous in-hand manipulation problem to enable the use of lookahead, we demonstrate its potential utility for many robotics applications.

\paragraph{RL Pretraining} 
To improve the sample efficiency of RL methods, research has explored augmenting demonstrations~\cite{rajeswaran2017DRL-demonstration,jena_augmenting_2020,radosavovic2021State-onlyImitation}, learning from priors~\cite{singh2020parrot}, and using model-based RL~\cite{nagabandi2020model-based-RL}. However, most of these behavior cloning approaches utilize human-collected trajectories as the expert policy to clone, limiting the amount and variety of data from which to learn. This work instead proposes using the optimal policies generated by DP on a discrete subset of the environment as the expert policy to clone. This obviates the need to commission a human study for data collection.


\section{Problem Definition}
\label{sec:problem}
The problem of interest is in-hand manipulation of a rigid body \emph{tool} through free space such that, once at the goal pose, it can apply a specified wrench on a given quasi-static object in the environment. We are motivated by the following example applications: using a wrench to tighten a nut, using a screwdriver to tighten a bolt, steadying a drill as it screws, or using a crow bar to pry open a box. We simplify this problem slightly and formalize it as:
\begin{itemize}
    \item {\bf Given:}
    \begin{itemize}
        \item An environment of static obstacles
        \item A rigid tool to manipulate
        \item A rigid multi-fingered robotic hand controllable via single DOF joint torques and a base $SE(3)$ free joint
        \item A target tool position $p_O^d$ and orientation $R_O^d$
        \item An external wrench $w_{ext}$ (i.e., force and torque) to be applied to the tool at that pose indefinitely
    \end{itemize}
    \item {\bf Find:} A sequence of joint torques such that
    \begin{itemize}
        \item The tool reaches its target pose without colliding with the environment or being dropped, and 
        \item The tool can maintain the target pose indefinitely under the applied $w_{ext}$.
    \end{itemize}
\end{itemize}
\subsection{Reduction to Grasp Sequence Planning}
The in-hand manipulation problem presented above is very broad. Our work uses a hierarchical planning and control framework~\cite{zarrin_hybrid_2022} to isolate the specific problem of grasp sequence planning. For each episode, a collision-free trajectory is planned for both the tool and the robot hand base through the environment to the target pose. This trajectory and the physical description of the tool (i.e. geometry, mass, moment of inertia, and friction coefficient) constitute the input provided to the grasp sequence planning algorithm at the start of each episode.

Low level continuous joint torque control is provided by the Contact Controller. This is a combination of inverse dynamics calculations, admittance control, and PID control that accepts both a reference tool pose and a reference grasp. It will then output joint torques to realize this pose and grasp given the physical state of the system.

Therefore, the general in-hand tool manipulation problem reduces to grasp sequence planning. At each time step, given the desired tool trajectory and system state, determine the reference grasp that needs to be sent to the Contact Controller.

\subsection{Grasp Sequence MDP Definition}
\label{sec:mdp}
We formulate the Grasp Sequence Planning problem as a finite-horizon Markov decision process (MDP) dependent on the Contact Controller~\cite{zarrin_hybrid_2022}.
\begin{itemize}
    \item {\bf State Space $\mathcal{S}$: }
    \begin{itemize}
        \item Robot joint positions $q$
        \item Robot base position $p_h$ and orientation $R_h$
        \item Tool pose $p_O, R_O$, plus linear and angular velocities $\dot{p_O}, \omega_O$
        \item Episode Input: desired tool trajectory $\{\widehat{p_O}(t), \widehat{R_O}(t)\}$, stopping time $T$, tool model, and its estimated inertial properties
    \end{itemize}
    \item {\bf Action Space $\mathcal{A}$: } Reference Grasp $\hat{G}$ for the Contact Controller
    \item {\bf Transition Function $\mathcal{S} \times \mathcal{A} \rightarrow \mathcal{S}$:} the composition of the Contact Controller and the environment's physical dynamics
    \item {\bf Reward $\mathcal{S} \times \mathcal{A} \rightarrow \R$ : } For $t < T$, return $R_{min}<<0$ if the tool or robot collides with the environment, and 0 otherwise. The reward for $t = T$ is described below.
\end{itemize}
At the end of the reference trajectory, i.e., at time $T$ when the tool should have reached its desired pose $\{\widehat{p_O}(T), \widehat{R_O}(T)\}$, we assume $w_{ext}$ is applied to the tool for an unknown time $T_{eval}$. Again, under any collision, the episode ends with the minimum reward $R_{min}$. Otherwise, the following cost is incurred:\begin{align}
\label{eq:cost}
    \int_{t = T}^{T + T_{eval}} \left[\twonormsq{p_O(t) - \widehat{p_O}(T)} + angle(R_O(t), \widehat{R_O}(T)) \right]dt
\end{align}
This mean-squared error calculation captures the ability of the grasp to maintain a constant pose under the desired wrench, which we assert is equivalent to applying that wrench to a static object at the given pose.

\section{Preliminaries}
\subsection{DP with Bellman Backups}
Consider any general MDP defined by a finite state space $\mathcal{S}$ with initial state $s_0$, a finite action space $\mathcal{A}$, some transition function $a_{t+1} = \mathcal{T}(s_t, a_t, w_t)$ (where $a_i \in \mathcal{A}$, $s_i \in \mathcal{S}$ and $w_i$ is a random variable drawn from a known distribution), some reward function $R(a_t, s_t, w_t)$, and a finite time horizon $t \in [0, T]$. DP \cite{bertsekas_dynamic_2012} can be used to find a policy $a_t = \pi(s_t)$ that maximizes the expected total reward $\sum_{t=0}^T\E_{w}\left[R(\pi(s_t), s_t, w)\right]$.

The solution relies on the \emph{principle of optimality}. If a hypothetical policy is optimal from some time $t$ until the end, then it contains within it the optimal policy from any intermediate time to the end. This principle allows the original MDP to be broken into simpler sub-problems by starting at the end and optimizing backwards.

One method for doing this is \textit{finite value iteration}, or the \textit{Bellman backup}, outlined in Algorithm \ref{alg:bellman}. Define the \emph{value}, or reward-to-go, of a state at a given time to be the sum of rewards that the optimal policy can obtain by starting at that state. We can write this value as a recursive function dependent on the value of the following states. For simplicity, we consider a deterministic reward and transition function.
\begin{align}
    V(s, t) &= \max_\pi\sum_{t' = t}^TR(\pi(s_t'), s_t') \nonumber \\
    &= \max_a [R(a, s_t) + V(s_{t+1} = \mathcal{T}(a, s_t))]
\end{align}
By beginning at the final time and working backwards until the start state, this algorithm can solve the initial problem in $O(T|\mathcal{S}||\mathcal{A}|)$ time. For very large or continuous state spaces, the set of reachable states can be determined by building a tree with all possible actions. The problem then reduces to an expectimax search over the tree with complexity $O(|\mathcal{A}|^T)$.

\begin{algorithm}[t!]
\caption{DP Using Bellman Backups}\label{alg:bellman}
\DontPrintSemicolon
\Input{States $\mathcal{S}$, Actions $\mathcal{A}$, Reward $R: \mathcal{S}\times\mathcal{A}\rightarrow\R$, Transition $\mathcal{T}: \mathcal{S} \times \mathcal{A} \rightarrow \mathcal{S}$, Horizon $T$, start state $s_0$}
\Output{Policy $\pi : \mathcal{S} \rightarrow \mathcal{A}$ and Value $V(s_0)$}
\BlankLine
$V \leftarrow array(|\mathcal{S}|\times T)$; $\pi \leftarrow array(|\mathcal{S}|\times T)$\;
$V[:, T-1] \leftarrow$ Terminal Reward for all $s \in \mathcal{S}$\;
\For{$t \leftarrow T-2, T-3, ..., 0$}{
    \For{$s \in \mathcal{S}$}{
        $V[s, t] \leftarrow \max_a R(s, a) + V[\mathcal{T}(s, a), t+1]$\;
        $\pi[s, t] \leftarrow \arg\max_a R(s, a) + V[\mathcal{T}(s, a), t+1]$\;
    }
}
\KwRet $\pi[s_0, 0]$, $V[s_0, 0]$\;
\end{algorithm}

\subsection{Online Lookahead and Rollout}
DP as described above has several limitations. It requires an accurate model of both the reward and transition functions, discrete state and action spaces, and the computation capacity to evaluate the Bellman backups for all possible states and actions. Additionally, the lack of any function approximation means that small perturbations in the reward model require DP to be completely re-run in order to recover the optimal policy. Therefore, solving the in-hand manipulation problem using DP is not feasible for real-world applications, where we expect a different context per episode and online variations in the transition and/or reward functions are inevitable.

In general, DRL solutions can overcome all these limitations using the function approximating power of neural networks. However, this approach incurs a trade-off between sample complexity and robustness to distributional shift. RL needs large amounts of valid data to approximate the transition and reward functions, and it can be thwarted by a test-time environment that differs substantially from the training environments. Ideally, we should be able to adapt an RL-trained policy to a new or perturbed environment without incurring  significant training costs or resorting to re-calculating the optimal policy from scratch with DP.

\begin{algorithm}[t!]
\caption{$l$-step Lookahead and $m$-step Rollout}\label{alg:lookahead}
\DontPrintSemicolon
\SetKwFunction{FRollout}{Rollout}
\SetKwFunction{FLookahead}{Lookahead}
\SetKwProg{Fn}{Function}{:}{}
\Input{States $\mathcal{S}$, Actions $\mathcal{A}$, Reward $R: \mathcal{S}\times\mathcal{A}\rightarrow\R$, Transition $\mathcal{T}: \mathcal{S} \times \mathcal{A} \rightarrow \mathcal{S}$, start state $s_0$}
\Input{Baseline Policy $\pi: \mathcal{S} \rightarrow \mathcal{A}$.\\ \emph{Optional:} Value Estimate $V: \mathcal{S} \rightarrow \R$}
\Input{Lookahead depth $l >= 1$, Rollout depth $m$}
\Output{Recommended action $a \in \mathcal{A}$}
\BlankLine
\Fn{\FRollout{State $s$}}{
        $v \leftarrow 0$\;
        \For{$t \in [0, m]$}{
            $v = v + R(s, \pi(s))$\;
            $s = \mathcal{T}(s, \pi(s))$\;
        }
        \KwRet $v + V(s)$\;
}
\Fn{\FLookahead{Current $l'$, State $s$}}{
        \uIf{$l' = 1$}{
            \KwRet $\arg\max_a$[$R(s, a) + $ Rollout($\mathcal{T}(s, a)$)]\;
        }
        \Else{
            \KwRet $\arg\max_a$[$R(s, a) + $ Lookahead($l'-1$, $\mathcal{T}(s, a)$)]\;
        }
}

\KwRet Lookahead($l$, $s_0$)\;
\tcc{Can return argmax or max as needed.}
\end{algorithm}

To address these limitations, we adopted the online $l$-step lookahead and $m$-step rollout algorithm \cite{bertsekas_lessons_2021} to solve the in-hand manipulation problem in a more data efficient and robust manner. Outlined in Algorithm \ref{alg:lookahead}, lookahead can be interpreted as performing DP with a time horizon $l$ (implemented as expectimax search). The terminal reward is determined by rolling out any policy (e.g., an RL-derived policy) for $m$ steps or until the end of the episode. In the former case, a trained value function (common in model-free RL such as PPO~\cite{ppo2017}) can be used as the terminal reward of the rollout. Importantly, the search tree and rollouts can be executed in a \emph{different} environment from that used to train the RL policy.

Lookahead provides tunable hyperparameters $l$ and $m$ to manage the trade-off between the high computational complexity of DP and any sub-optimality present in the underlying RL model in the novel environment.

\section{Scheme and Surrogate Model}
\label{sec:environment}
We propose to solve the MDP outlined in Section \ref{sec:mdp} with a trained RL policy augmented at test time (i.e., online) with the DP-based procedures described above. The actual lookahead and rollout occurs by alternating between the RL policy (given a state, output a grasp) and a surrogate environment model (given a grasp, predict the next state) implemented in OpenAI's Gym~\cite{brockman2016openai}. This model is populated with the episode input (i.e., desired trajectory and description of the robot and tool) at test time and utilizes discretization and a shaped dense reward function that are relatively quick to compute. Even given these approximations, we hypothesize that the result is an augmented RL policy that can adapt to previously unseen episodes without additional training.

\subsection{Discrete Action Space}
We rely on the assumption that each link between the robot and the contactable portion of the tool is convex so we can guarantee that each link makes contact with the tool at only one point. From there, the complete set of possible reference tool grasps $G$ is discretized by defining a finite countable set of contacts (consisting of one point on the robot and one on the tool) for each robot link, including a ``null'' contact that is the removal of the given link from the tool.

Due to the current structure of the Contact Controller, this assumption is subject to the restriction that only one robot link can be modified (either added, removed, or slid) at a time. Therefore, in addition to the null action that maintains the current reference grasp, an action consists of selecting which link to modify and its new reference contact point.

\subsection{Shaped Reward Function}
\label{sec:reward}
The reward function is constructed to be correlated with the odds of grasp failure (i.e., the tool falling) without full physical simulation by splitting kinematics and dynamics calculations into \emph{IK Error} and \emph{Wrench Error} respectively.

\paragraph{IK Error} IK is performed iteratively by collision-avoiding gradient descent from some fixed starting joint configuration, targeting tool contact points $p_O$. The result is a new joint configuration and a set of effective contact points $p_{R}$ on the link in the world reference frame. In this context, IK Error is defined as $\max_{all\ links}\norm{p_R - p_O}$, i.e., the worst error between the target contact point on the link and the target contact point on the tool.

\paragraph{Wrench Error} This is the solution to the following constrained optimization problem: find the set of contact forces within the friction cone (or on the edge in case of sliding contact) that minimizes the difference between the net wrench on the tool and the desired net wrench. An additional term in this metric penalizes the sum of the square norm of each contact force, which in turn penalizes grasps with few links and high contact forces.

Thus, the final reward function is constructed as follows:
\begin{itemize}
    \item $R_{min}$ if either IK fails with collision (including self-collision) or any \emph{Wrench Error} exceeds a falling threshold; otherwise:
    \item \emph{Wrench Error} against gravity and desired tool motion after realizing the new grasp at the current tool position with IK $+$
    \item \emph{IK Error} realizing the new grasp at the desired tool position $+$
    \item \emph{Wrench Error} against just gravity after this second IK calculation $+$
    \item \emph{Wrench Error} against $w_{ext}$ at end time $T$ $+$
    \item A small penalty for redundant actions (i.e., commanding a non-null action that does not result in a changed grasp)
\end{itemize}

State transition is performed simultaneously with reward calculation. The output of the IK procedures sets the joint positions of the robot hand that are used to determine the approximate system state without full physical simulation.

\subsection{Discrete State Space}
\label{sec:state}
The discrete grasp command and the trajectory waypoints themselves do not make the state space discrete. No restrictions are imposed on the observed tool position or joint states, which remain continuous at test time. In fact, a discrete state space is not necessary for lookahead, rollout, or RL training (which consists of rollouts to collect data for policy iteration).

However, discretization is necessary to tractably use Bellman backups to completely ``solve'' an environment, and we use it for DP pretraining and some of our baseline policies. We do this by making the following assumptions:
\begin{itemize}
    \item The tool perfectly adheres to the reference trajectory. Therefore, all possible current tool positions come from a discrete set of waypoints.
    \item A given grasp is always realized as well as possible, as defined by running the IK procedure  in Section \ref{sec:reward} with a fixed initial joint configuration. This defines a unique joint configuration for a given discrete tool position and reference grasp.
\end{itemize}
From these assumptions as well as a given reference trajectory, grasp, and discrete time step, we can determine a unique \emph{discrete state}. For Bellman-backup-based DP, we can operate entirely on these discrete states.

\begin{figure}[t!]
         \centering
         \includegraphics[width=\linewidth]{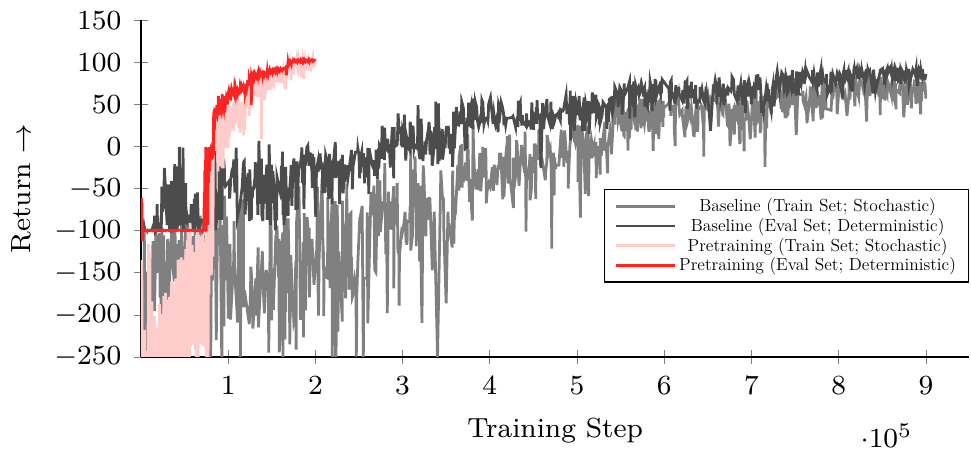}
         \caption{Training curves for the RL model with and without DP pretraining. Lighter lines are the performance of the stochastic policy on the training set. Darker lines are the performance of the deterministic policy on the validation set, used to determine the training stop time. With DP pretraining, the RL policy can achieve the same return in four times fewer training iterations.}
         \label{fig:pretrain}
         \vspace{-0.3cm}
\end{figure}
\subsection{DP Pretraining Procedure}
Deep RL algorithms that follow the ``actor-critic'' model involve estimating two functions: given any state, the \emph{policy function} recommends an action, and the \emph{value function} estimates the value (i.e. sum of future rewards). By performing DP to completely solve the discretized environment and recording intermediate results, we can determine the optimal action and the theoretical value of each state. In effect, we generate a behavior cloning dataset that can be used to directly train these networks with any supervised learning algorithm.

\begin{figure*}[t!]
     \centering
     \begin{subfigure}[b]{0.49\textwidth}
         \centering
         \includegraphics[width=\textwidth]{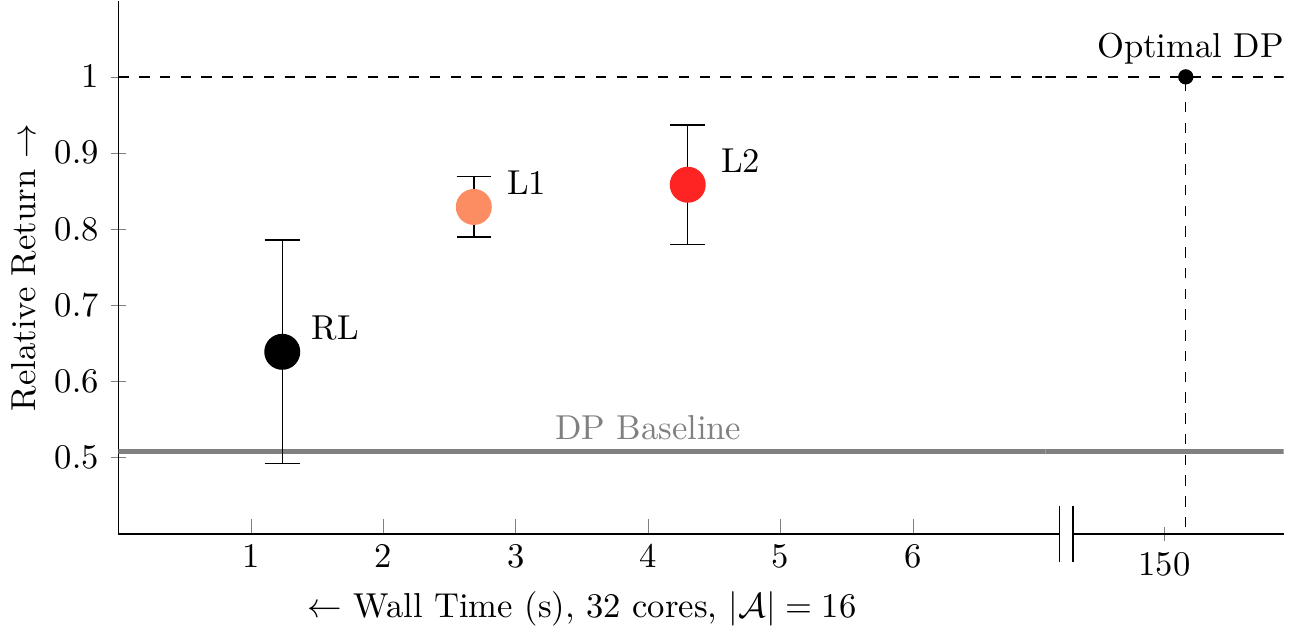}
         \caption{Average return (relative to the optimal policy) of RL, 1-step Lookahead (L1), and 2-step Lookahead (L2), as evaluated within the surrogate environment model. DP Baseline is the optimal policy across previously seen trajectories. Error bars represent 95\% confidence. Lookahead significantly outperformed RL $(P < 0.05)$ with only a few seconds more computation time on a 32-core machine.}
         \label{fig:offline}
     \end{subfigure}
     \hfill
     \begin{subfigure}[b]{0.49\textwidth}
         \centering
         \includegraphics[width=\textwidth]{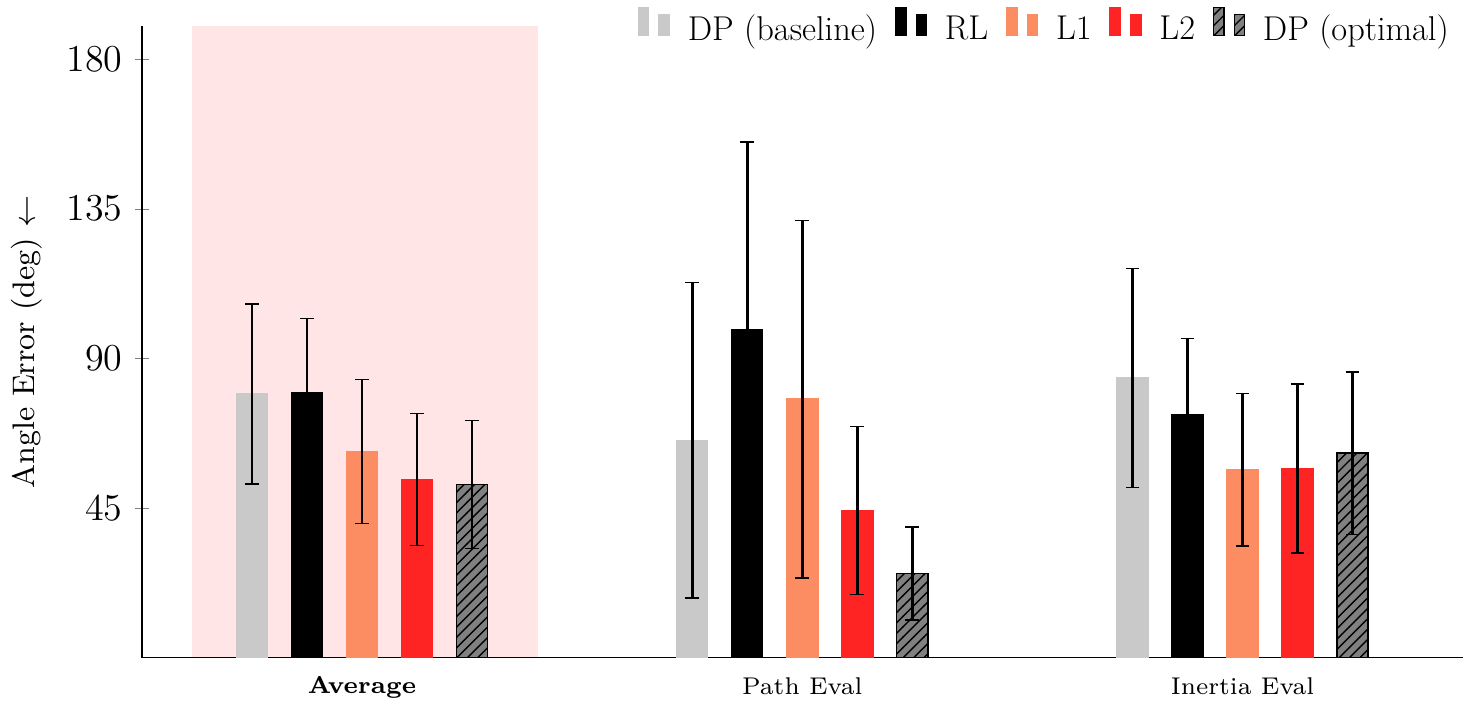}
         \caption{Average angle error of the tool while applying desired final torque. DP (optimal) was solved on each evaluation trajectory. DP (baseline) was solved only on the training trajectories. Error bars represent 95\% confidence. These results suggest that our surrogate model is correlated with the simulation environment and that lookahead (L1, L2) can approach DP optimal performance. Note that DP ``optimal'' is no longer guaranteed to be optimal in the full simulation, as it is solved on the surrogate model.}
         \label{fig:sim_results}
     \end{subfigure}
     \caption{Results on the 27 evaluation episodes for (\emph{left}) surrogate environment model and (\emph{right}) physical simulation.}
     \vspace{-0.56cm}
\end{figure*}

This procedure yields two key benefits. First, while the DP solution is not optimal for the non-discrete setting, the resulting policy is likely closer to optimal than just using random initial network weights. This is supported empirically by the good performance of the DP solution within the non-discrete simulation (see \figref{fig:sim_results}). Second, running cross-validation on the DP-generated dataset can allow for easy hyperparameter tuning. For example, in our case, we used DP pretraining to tune our actor and critic network sizes and select ReLu over Tanh as the activation function. As a result, we could train our RL models to the same validation-set performance with 4x fewer training steps, as shown in \figref{fig:pretrain}.


\section{Evaluation}
\label{sec:eval}
We evaluated our surrogate model and online augmentation scheme in simulation with Honda's 4-fingered robotic hand \cite{hasegawa2022MFH} and a rigid 0.35kg ($m_0$) adjustable wrench, as shown on the left side of \figref{fig:system}. The number of possible contact points per link were defined as follows: thumb distal (3 points), index distal (3 points), index medial (2 points), middle distal (4 points), middle medial (2 points), ring distal (2 points).

The product of these contact points yields 288 discrete states and reference grasps, and the sum yields 16 possible actions (i.e.,  moving only one finger at a time). All wall-clock timing calculations were done on a 32-core Intel(R) Xeon(R) Silver 4216 clocked at 2.10GHz.
\begin{table}[t!]
\centering
\begin{tabular}{c c c c c} 
 \toprule
 DP (base) & RL & $l=1$ & $l=2$ & DP (opt) \\ 
\midrule
1.34s & 1.23s & 2.68s & 4.30s & 151.79s \\ 
\bottomrule
\end{tabular}
\caption{Online Wall-clock Time (s) Per Unseen Trajectory}
\label{tab:cpu}
\end{table}
\subsection{Experiment Design}
We evaluated this approach on a series of tool trajectories. The \emph{nominal reference trajectory} started the wrench on a planar surface at the origin, i.e., 15cm below the palm of the robot. The first waypoint was at $(0, 0, 0.014)$, and 15 more waypoints moved the wrench to $(0, 0, 0.06)$ according to the reference velocity and acceleration before applying a positive 1Nm torque about the Z-axis relative to the center of mass.
From the nominal trajectory, we varied the height of both the initial and final waypoints, torque direction, tool mass, and center-of-mass y-offset to create a training set and two evaluation sets. The training set included 9 reference trajectories: the nominal trajectory, the nominal trajectory with a negative final torque, and height variations from -6mm to +8mm in increments of 2mm, where negative height variants (-6, -4, and -2mm) experienced a negative final torque, and vice versa. The first evaluation set measured robustness to path deviation and included 7 trajectories: height variation from -6mm to +8mm in increments of 2mm (excluding the nominal 0mm), where the \emph{positive} height variants (+2, +4, +6, and +8mm) experienced a negative final torque, and vice versa. The second evaluation set measured robustness to mass and center-of-mass deviation and included 20 trajectories: masses varied from $0.5m_0$ to $2m_0$ in increments of $0.5m_0$, and the center-of-mass y-offset varied from $-2cm$ to $2cm$ in increments of $1cm$.

\subsection{Algorithms}
Table \ref{tab:cpu} shows the online computation time (i.e., excluding training time) for each algorithm. For reference, each trajectory took approximately 2 minutes to run in simulation at 1x (i.e. real-world) speed.
\paragraph{DP Baselines} For each trajectory, the reward associated with each discrete state-action-time triplet was recorded, for a total of approximately 73k (288 states $\times$ 16 actions $\times$ 16 time steps) executions of \texttt{env.step()}. Unreachable states (e.g., requiring moving more than one finger at a time) were assigned a reward of $-\infty$. We then ran DP on this data to determine the optimal grasp sequence for that trajectory under discretization. We refer to the DP solution on the nominal trajectory as the ``DP Baseline'' and on each evaluation trajectory as ``Optimal DP.'' In Table \ref{tab:cpu} and \figref{fig:offline}, DP solution computation time counts against ``Optimal DP,'' which must be solved during test time after the evaluation trajectory is revealed, but not against ``DP Baseline,'' which is solved at training time. DP benefits from a 32x speedup from parallelization on our system.
\paragraph{RL Baseline} We used the PPO Clip Agent from TFAgents \cite{TFAgents} 3-layer feed-forward policy and value networks and an Adam optimizer with a learning rate decreasing piece-wise-linearly. We generated a gym training environment that cycled through all training trajectories, and RL states were not limited to the discrete states of DP.
\paragraph{Lookahead Algorithms} We investigated the 1- and 2-step lookahead algorithms (``L1'' and ``L2,'' respectively). Since the episode lengths were short (at 16 time steps maximum), we ran our RL-baseline policy all the way to the end of the episode for the rollout portion of the algorithm (i.e.,  $m = T - t - l\;\ \forall t$). Each rollout can run in parallel, yielding a 16x ($|\mathcal{A}|$) speedup for L1 and a full 32x speedup for L2 (number of cores).

\subsection{Results}
\subsubsection{Surrogate Model}
We evaluated all trained algorithms on the evaluation trajectories in our surrogate environment model. \figref{fig:offline} presents return results, with significance determined by running a 1-way ANOVA with standard weights and samples correlated (by trajectory). All algorithms significantly outperformed the DP Baseline, and lookahead significantly outperformed RL $(P < 0.05)$. While we only tested up to L2, the results are consistent with the idea that lookahead experiences diminishing returns with larger trees \cite{bertsekas_lessons_2021}. Therefore, most of the improvement comes from first step (L1), requiring a relatively small amount of additional computation.

\subsubsection{Simulation}
We evaluated the applicability of the surrogate model to real manipulation in a full physics simulation using a rigid-body physics engine \cite{zarrin_hybrid_2022}. The original MDP reward (Equation \ref{eq:cost}) is implemented by manually applying the external wrench $w_{ext}$ at the end of the trajectory and measuring position and angle error. Since $w_{ext}$ consisted of only a torque term, there was little difference in the position term. Therefore, \figref{fig:sim_results} shows only the angle error term. If the tool collided with the floor, the maximum error of $180$ degrees was assigned.

The fact that ``Optimal DP'' performed best across the evaluation set suggests that the shaped reward of our surrogate model is correlated with the desired MDP reward function. Additionally, these results are consistent with the previous ones in that the lookahead algorithm generally outperformed baseline DP and RL, approaching the optimal DP performance.
\section{Conclusion}
We proposed a data-efficient and robust approach for solving the in-hand tool manipulation problem and evaluated our results in simulation using a dexterous hand that performs tool-use under modeling uncertainties and task variations. Results show that, given even a discrete approximate environment model, DP can be used to augment both the training and execution of RL policies with minimal online computational overhead. We believe that these insights can be applicable to any problem where such a model can be constructed.

Future work opportunities include action space expansion to enable performing additional manipulation primitives. Though a larger or continuous action space will make the lookahead tree more complex to evaluate, we could remedy this using a heuristic-based search technique. Implementing the proposed algorithms on the hardware is underway, and we will then evaluate results on the real system.

\newpage
\bibliographystyle{IEEEtran}
\bibliography{references}
\end{document}